\pdfoutput=1

\documentclass[11pt]{article}

\usepackage[preprint]{acl}

\usepackage{times}
\usepackage{latexsym}

\usepackage[T1]{fontenc}

\usepackage[utf8]{inputenc}

\usepackage{microtype}

\usepackage{inconsolata}

\usepackage{graphicx}

\usepackage{multirow}

\title{PoAct: Policy and Action Dual-Control Agent for Generalized Applications}

\author{
 \textbf{Guozhi Yuan\textsuperscript{3}\footnotemark[1]},
 \textbf{Youfeng Liu\textsuperscript{1}},
 \textbf{Jingli Yang\textsuperscript{1}},
 \textbf{Wei Jia\textsuperscript{1}},
 \textbf{Kai Lin\textsuperscript{2}},
\\
 \textbf{Yansong Gao\textsuperscript{4}\footnotemark[1]},
 \textbf{Shan He\textsuperscript{5}\footnotemark[1]},
 \textbf{Zilin Ding\textsuperscript{2}\footnotemark[1]},
 \textbf{Haitao Li\textsuperscript{4}\footnotemark[1]}
\\
 \textsuperscript{1}Zhipu AI,
 \textsuperscript{2}Amarcredit,
 \textsuperscript{3}Central South University,
 \textsuperscript{4}Tsinghua University,
 \textsuperscript{5}Beihang University,
\\
 \small{
    \href{mailto:youfeng.liu@zhipuai.cn}{youfeng.liu@zhipuai.cn} 
 }
}

\begin{document}
\maketitle

\renewcommand{\thefootnote}{\fnsymbol{footnote}} 
\footnotetext[1]{Work was done when interned at Zhipu AI.} 

\begin{abstract}
Based on their superior comprehension and reasoning capabilities, Large Language Model (LLM) driven agent frameworks have achieved significant success in numerous complex reasoning tasks. ReAct-like agents can solve various intricate problems step-by-step through progressive planning and tool calls, iteratively optimizing new steps based on environmental feedback. However, as the planning capabilities of LLMs improve, the actions invoked by tool calls in ReAct-like frameworks often misalign with complex planning and challenging data organization. Code Action addresses these issues while also introducing the challenges of a more complex action space and more difficult action organization. To leverage Code Action and tackle the challenges of its complexity, this paper proposes \textbf{Po}licy and \textbf{Act}ion Dual-Control Agent (PoAct) for generalized applications. The aim is to achieve higher-quality code actions and more accurate reasoning paths by dynamically switching reasoning policies and modifying the action space. Experimental results on the Agent Benchmark for both legal and generic scenarios demonstrate the superior reasoning capabilities and reduced token consumption of our approach in complex tasks. On the LegalAgentBench, our method shows a 20 percent improvement over the baseline while requiring fewer tokens. We conducted experiments and analyses on the GPT-4o and GLM-4 series models, demonstrating the significant potential and scalability of our approach to solve complex problems.
\end{abstract}
%

\section{Introduction}

With the rise of Large Language Models (LLMs), knowledge-driven frameworks for agents have gradually become mainstream. LLMs are increasingly capable of handling complex tasks due to their powerful reasoning abilities and extensive knowledge bases. The ReAct framework \citep{Yao2022ReActSR}, a significant advancement in this technology, offers substantial advantages. ReAct-like agents, by integrating reasoning and action, enable the decomposition of complex tasks into multiple manageable subtasks, facilitating a gradual resolution of the process\citep{Wang2023VoyagerAO,Zhu2023GhostIT}. ReAct not only enhances the execution efficiency of agents in complex tasks but also ensures that they can effectively address intricate problems by receiving environmental feedback and continuously adjusting and refining their policies during task execution. Furthermore, ReAct provides flexible reasoning and action capabilities for agents, demonstrating the benefits of disassembly and step-by-step execution in solving complex tasks, thereby establishing itself as one of the cornerstones of agent solutions.

\begin{figure}[t]
  \includegraphics[width=0.5\textwidth]{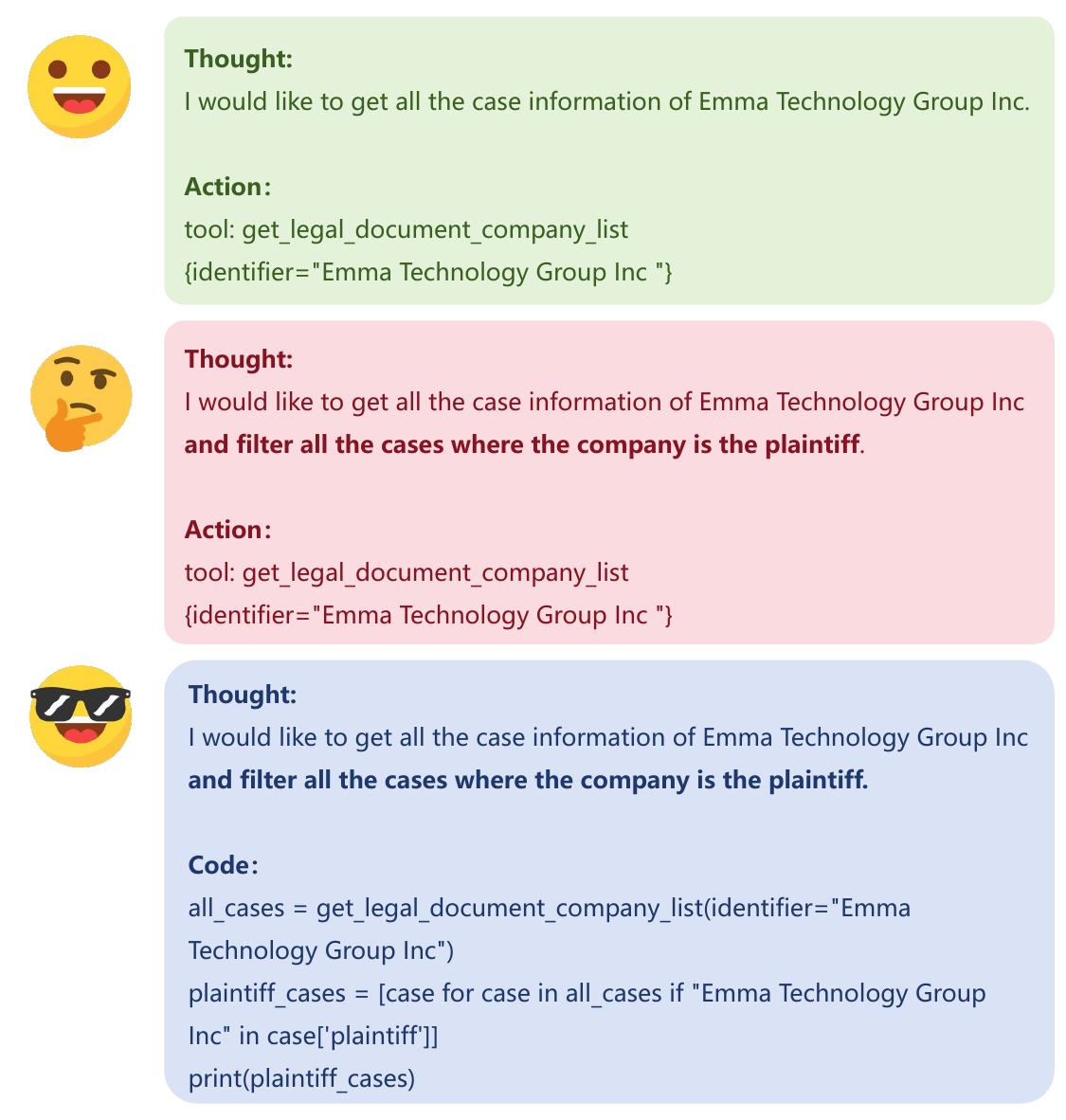} 
  \caption{The figure illustrates the mismatch between the single-step action space and the planning capabilities of LLMs. Code actions help mitigate this problem to some extent.}
  \label{fig:rca}
\end{figure}

The ReAct-like framework excels in task disassembly and step-by-step execution. However, it has limitations when addressing complex problems. While LLMs possess advanced planning and reasoning capabilities \citep{Zhao2023ASO}, the simple tool calls of ReAct struggle to align with the intricate planning of LLMs in certain steps. As illustrated in Figure \ref{fig:example}, when the complexity of LLMs' planning increases, a single tool call often fails to effectively match complex planning and manage structured data. Furthermore, the inability of a single tool call to handle large-scale textual data or non-textual modal data restricts the data exchange and execution efficiency of the Agent. To mitigate these challenges, CodeAgent \citep{Zhang2024CodeAgentEC} introduces Code Action, which allows for the handling of complex computational tasks through the flexibility of code, thereby significantly enhancing the efficiency of various state data exchanges. Additionally, leveraging the coding capabilities of LLMs enables agents to efficiently organize tools and Python packages, greatly expanding the action space of agents and improving the alignment between planning and actions.

Through the integration of ReAct and Code Action, the ReAct Code Agent (RCA) can execute complex and diverse action spaces while gradually optimizing the planning of each step \citep{wolf-etal-2020-transformers}. However, this advancement also introduces new challenges. Code Action significantly increases the complexity of the action space and complicates the organization of actions compared to single tool calls. During multiple rounds of planning and reasoning, the RCA must select the most appropriate tools from a large number of tools and organize high-quality Code Actions, which presents a considerable challenge for agents. Consequently, determining how to execute high-quality and accurate Code Actions within a complex reasoning process has become an urgent issue that needs to be addressed.
%

To address this challenge, we draw inspiration from human teamwork. An efficiently functioning team often requires individuals to assume different roles that cooperate in solving a task, with each role focusing solely on the impact of specific responsibilities rather than the details of other steps. Inspired by this paradigm, this paper proposes Policy and Action Dual-Control Agent (PoAct) for generalized applications. The PoAct can dynamically adjust its action space and reasoning policy according to various reasoning steps. Specifically, PoAct introduces a Policy Controller that emphasizes high-quality planning and coding through the perspectives of expert roles in stages such as Planning, Thought, and Code Action. The Policy Controller ensures that PoAct concentrates on the current reasoning step rather than relying on a single policy throughout the process. Additionally, the shared conversation history and action space help minimize information loss. To address the challenges posed by complex action spaces, this paper also designs a Code Action Controller, which includes a RAG Selector for dynamically selecting the most appropriate action space among different expert perspectives, and an Action Reviewer for evaluating and controlling the PoAct's reasoning paths, effectively preventing deviations from the expected actions.

We evaluated PoAct on multiple task datasets and conducted experiments using both commercial and open-source large models. The experimental results demonstrate that PoAct exhibits outstanding performance across various tasks, showcasing high generalizability and scalability. Our primary contributions include:

\begin{itemize}
    \item We propose the Policy Controller module, which is designed to dynamically switch reasoning policies, allowing PoAct to concentrate on solving specific reasoning steps. This module utilizes various expert perspectives to collaborate, resulting in a higher quality and more efficient reasoning process.
    \item We introduce the Action Controller module, designed to dynamically manage the action space of various reasoning steps and to review the reasoning path, thereby ensuring a highly stable reasoning process.
    \item We evaluate PoAct's capabilities on multiple agent benchmarks, and the experimental results from different tasks demonstrate PoAct's robust reasoning abilities and scalability.
\end{itemize}

\begin{figure*}[ht]
  \includegraphics[width=1\textwidth]{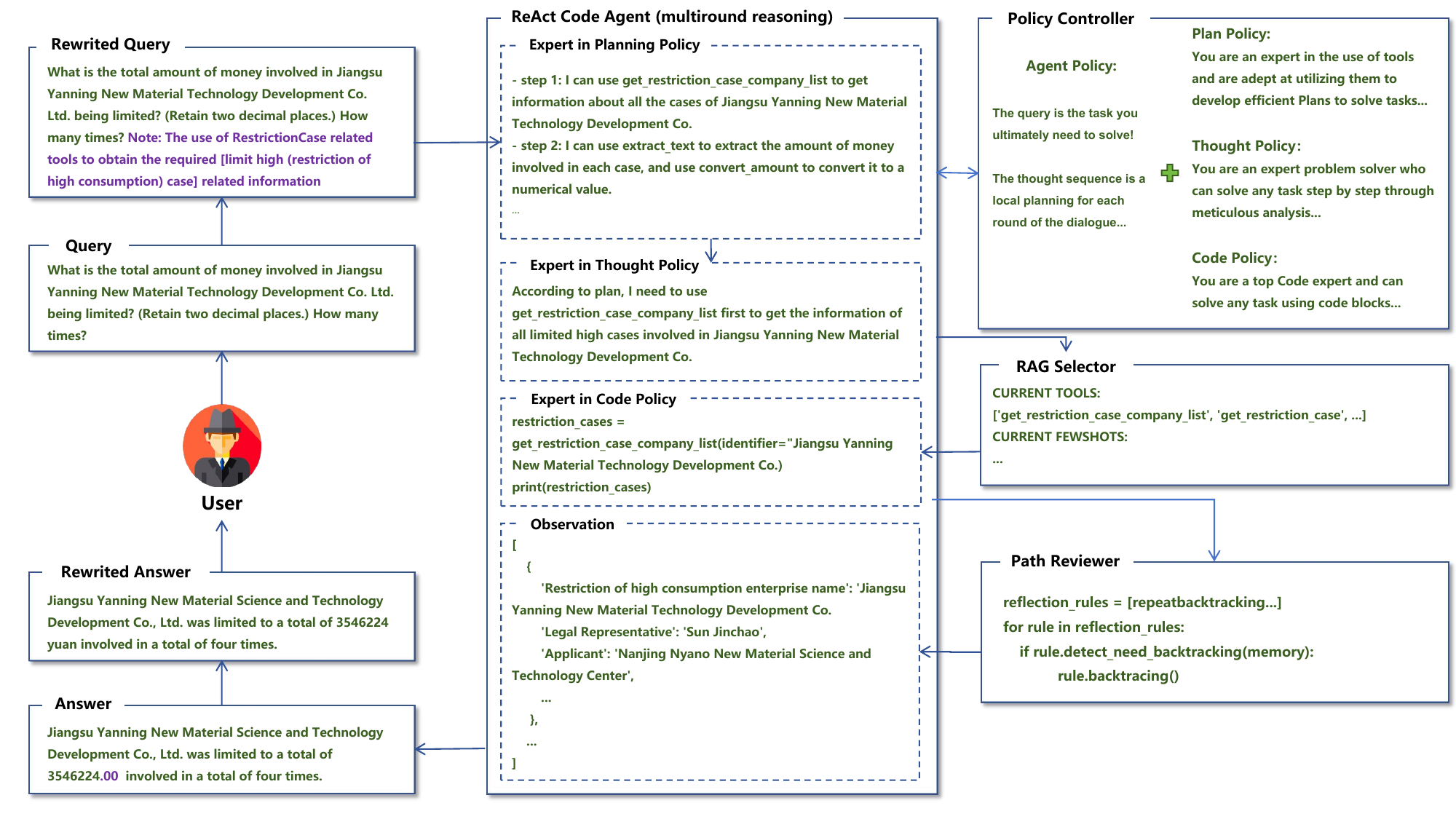} 
  \caption{The framework diagram of PoAct illustrates how PoAct adjusts its reasoning policy based on the reasoning step via the Policy Controller. It employs the Action Controller to dynamically switch the visible tools and few-shot examples, while also evaluating anomalous reasoning paths.}
  \label{fig:PoAct}
\end{figure*}

\section{Related Works}




\subsection{Large Language Model Based Agent}

Large Language Models (LLMs) demonstrate powerful reasoning and complex task-processing capabilities \citep{Radford2019LanguageMA}. LLMs such as GPT-3, GPT-4, and PALM perform well in multi-round dialogue comprehension and response generation. \citep{Floridi2020GPT3IN,Thoppilan2022LaMDALM,Achiam2023GPT4TR}. LLMs are equipped with contextual learning, instruction following, and chain-of-thought reasoning, which facilitate the transition from rule-based to knowledge-driven approaches for agents \citep{Ouyang2022TrainingLM,Wei2021FinetunedLM,Wei2022ChainOT}. AgentGym \citep{Xi2024AgentGymEL} demonstrates the capability of a single LLM agent to adapt across multiple tasks and environments, gradually evolving from imitative learning to interactive learning and exhibiting human-like learning behaviors. AssistGPT \citep{Gao2023AssistGPTAG} enhances agents' integration with various tools through planning, verification, execution, and reasoning methods, thereby improving their adaptability in diverse application scenarios. CodeAgent \citep{Zhang2024CodeAgentEC} combines code generation and execution, dynamically switching tasks through a code pattern controller, demonstrating higher flexibility and multi-tool calling capability. Nevertheless, LLM-based agents often face limitations in reasoning capabilities when addressing cross-domain tasks, and their restricted action space hinders autonomous decision-making and task execution in complex environments.

\subsection{Code Enhanced Tool Reasoning}
Improving reasoning efficiency and flexibility presents a significant challenge in real-world applications. Even well-designed multi-agent frameworks struggle to surpass the inherent reasoning capabilities of LLMs, despite careful instruction and demonstration on collaborative tasks \citep{Wang2024RethinkingTB}. Numerous studies have employed text or JSON formats to generate actions for tool calling \citep{Park2023GenerativeAI,Qin2023ToolLLMFL,Wang2023VoyagerAO}, but these methods exhibit limitations in operational space and flexibility. Recent research has begun to investigate the use of LLMs for code generation to enhance manipulation capabilities. For instance, Voyager \citep{Wang2023VoyagerAO} applies code generation within the gaming domain, while Code4Struct \citep{Wang2022Code4StructCG} utilizes code generation for structured prediction tasks. However, these approaches typically generate code in a single round, making dynamic adaptation challenging. TaskWeaver \citep{Qiao2023TaskWeaverAC} integrates code into the agent manipulation space but remains at the conceptualization level. Recent studies have employed a code ReAct approach to generate Python code via a Python interpreter and a collection of necessary packages, allowing for dynamic tuning of operations and demonstrating limited capabilities for self-improvement and environmental interaction \citep{wolf-etal-2020-transformers}. This approach not only manages and adapts tool usage but also responds dynamically to environmental changes, facilitating efficient task completion. However, due to the complexity of code, efficiently generating and executing correct, high-quality code remains a challenge.
%

\section{PoAct: Policy and Action Dual-Control Agent}

In this section, we will provide a detailed introduction to the PoAct and its components. As illustrated in Figure \ref{fig:PoAct}, the PoAct framework is grounded in the reasoning paradigm of the ReAct Code Agent. To enable PoAct to concentrate on specific reasoning steps, the Policy Controller dynamically adjusts the step policies for different reasoning phases, allowing PoAct to focus solely on the context and reasoning intricacies of particular steps. Additionally, the Action Controller optimizes the action space and evaluates abnormal code actions using a RAG Selector and an Action Reviewer.

\subsection{ReAct Code Agent}

\begin{figure}[t]
  \includegraphics[width=0.5\textwidth]{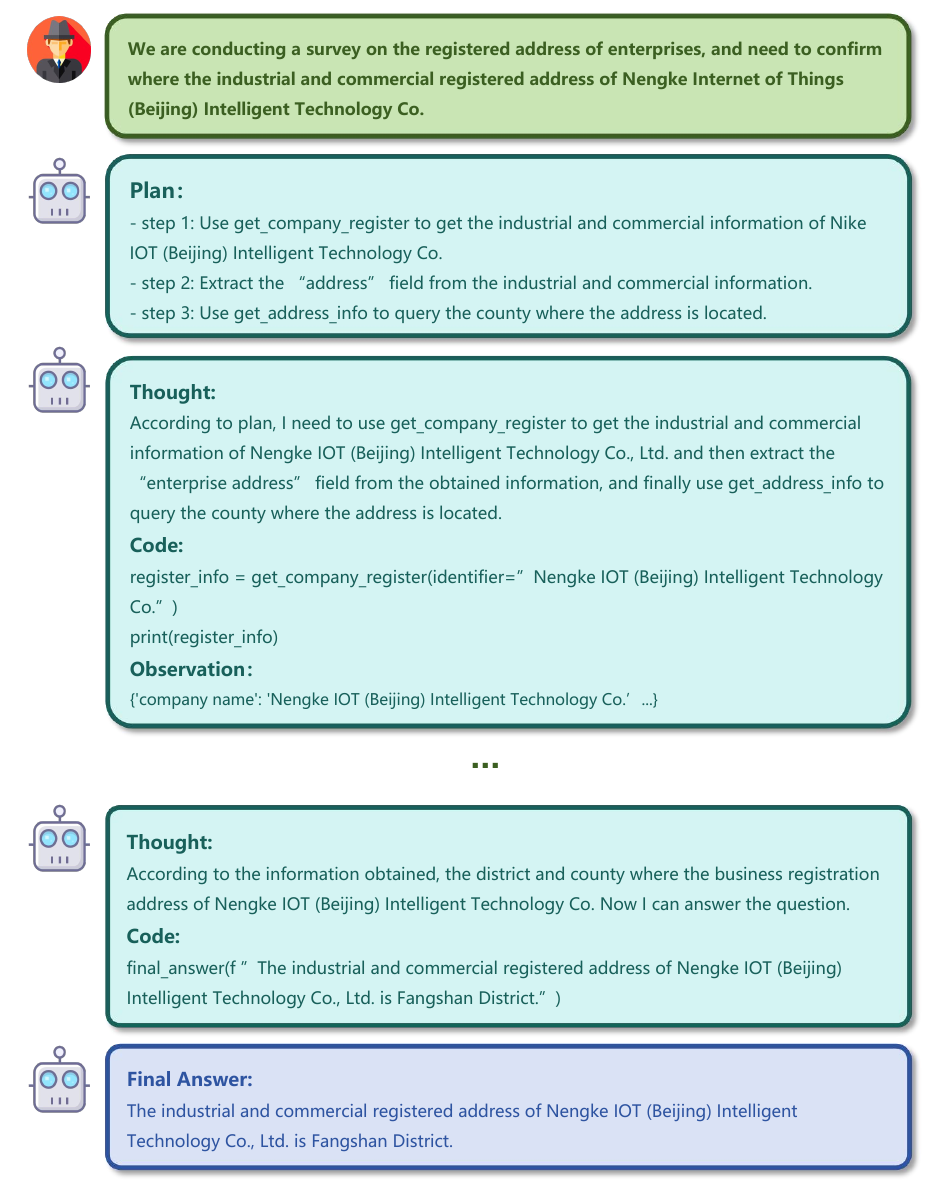} 
  \caption{This diagram illustrates how PoAct addresses user queries using the ReAct Code paradigm.}
  \label{fig:example}
\end{figure}

PoAct employs the ReAct Code paradigm, which integrates the ReAct Agent and Code Agent. This paradigm effectively combines the reasoning capabilities of LLMs with function-calling policies, similar to the ReAct Agent, while also enabling more complex actions than simple function calls through a code action space. As illustrated in Figure \ref{fig:example}, after receiving a task, PoAct first executes the planning step to decompose the complex task into subtasks. It then engages in step-by-step planning and updates the global plan when the task cannot be resolved immediately. PoAct initiates a multi-step reasoning process based on this planning, generating local planning a thought for each step to guide the generation and execution of code actions. The process is informed by the results of these actions, referred to as observations. This iterative approach continues, generating subsequent rounds of reasoning based on the observations, until PoAct successfully deduces the answer to the user's task. By leveraging code actions, PoAct can navigate highly complex action spaces and manage all states flexibly using variables.

\subsection{Policy Controller}

The ReAct Code paradigm enhances the capability of agents to navigate complex environments and increases the challenge of organizing accurate and high-quality code actions. Existing frameworks for agents typically depend on a single policy to address the entire reasoning process, which can lead to the agent prioritizing the maintenance of the reasoning process while neglecting the details and implications of individual reasoning steps. To mitigate this issue, we employ a Policy Controller to manage the PoAct in switching between different reasoning policies at various stages of the reasoning process (Planning, Thinking, Coding) to concentrate on executing specific steps. Specifically, the Policy Controller dynamically integrates the system prompt, including an agent policy prompt that aids the agent in comprehending the dialogue history, and a step policy prompt that guides the agent to focus on a particular reasoning step. This section describes in detail how Policy Controller switches different reasoning policies for different reasoning steps.

\subsubsection{Agent Policy Prompt}

The agent policy prompt primarily establishes a global policy for the PoAct, aiming to clarify the structure of the dialogue history. As illustrated in Figure \ref{fig:PoAct}, we define the meanings of query, thought, code and observation in agent policy prompt. This clarification helps PoAct maintain a comprehensive reasoning process and prevents confusion when switching between policies.

\begin{figure*}[ht]
  \includegraphics[width=1\textwidth]{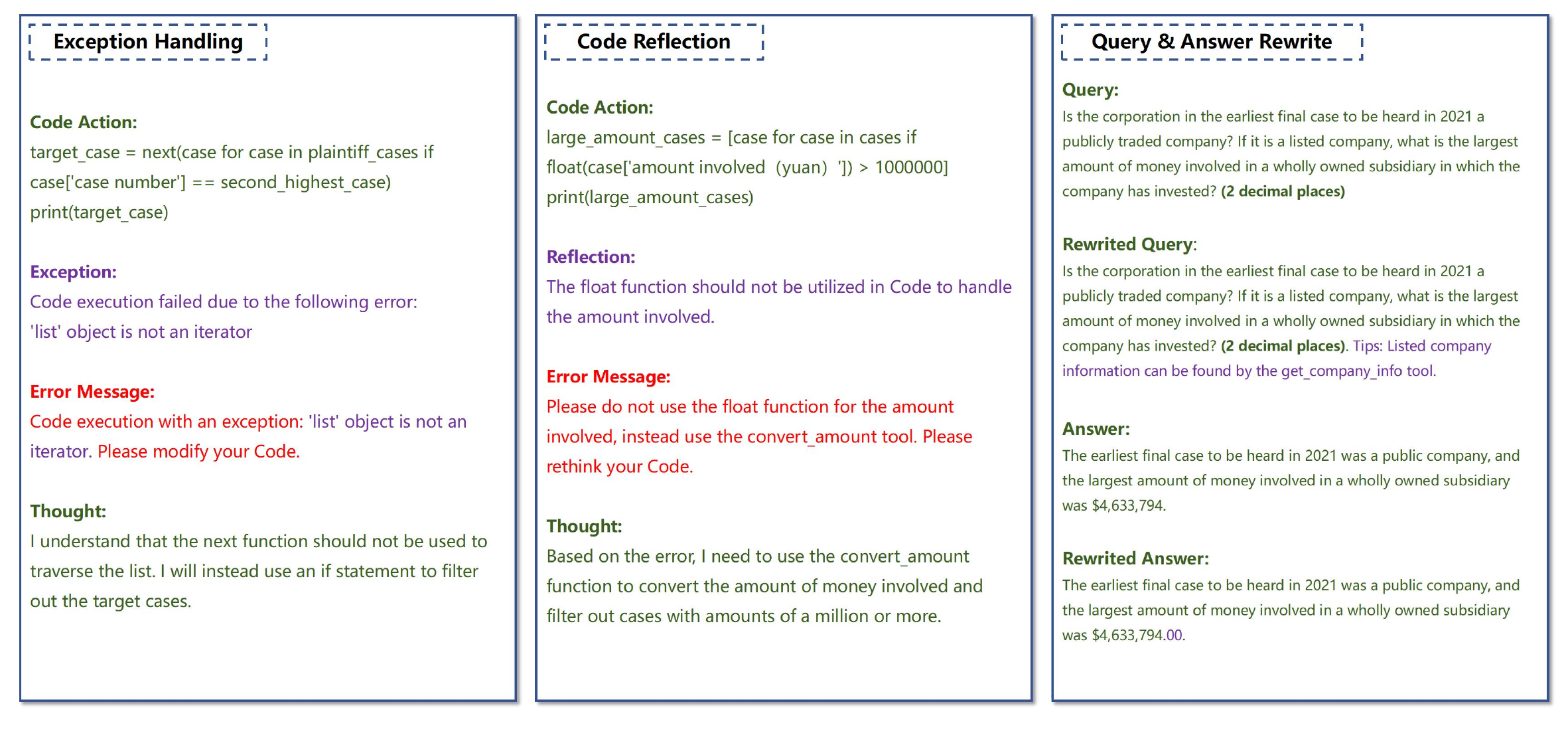} 
  \caption{Path Reviewer avoids PoAct from entering anomalous reasoning paths through three tasks.}
  \label{fig:path reviewer}
\end{figure*}

\subsubsection{Step Policy Prompt}

Step policy prompts primarily allow PoAct to transition into specific roles during various reasoning steps, allowing them to concentrate on the details and quality of the current step. Policies that emphasize particular reasoning steps can enhance the quality of each step while minimizing distractions from the demands of other steps. Furthermore, the division of a single policy into prompts that target specific reasoning steps reduces the time and effort required for prompt encoding. In the following, we will briefly outline the focus of the different policies.

\textbf{Plan Policy} requires PoAct to focus on decomposing complex tasks into subtasks and providing a comprehensive plan for each subtask to guide subsequent reasoning steps. PoAct must adopt the perspective of a planning expert, who excels in global planning, to develop a complete and executable plan utilizing all available tools and packages during the execution of plan steps. The planning expert concentrates solely on breaking down queries that can be parallelized and does not require detailed knowledge of the specific code execution for the subsequent steps. The planning phase typically occurs before multiple rounds of reasoning and involves revising the plan based on the dialogue history as challenges arise during the reasoning process.

\textbf{Thought Policy} mandates that the PoAct to develop a local plan for the current round, drawing on the global plan and conversation history, which have informed the generation of the next code action. To effectively execute the thought step and create a local plan, PoAct requires the insights of a thought expert skilled in multi-step problem solving. This expert will concentrate solely on formulating an appropriate local plan by analyzing past conversation history before directing the subsequent steps necessary to obtain the required observations.

\textbf{Code Policy} requires PoAct to generate code actions for the round based on local planning. PoAct must adopt the perspective of a professional code expert while executing the Code steps to effectively translate the objectives of the round into high-quality and robust code actions. The code expert views tools as functions and emphasizes the importance of utilizing the appropriate functions to complete the code logic for the round. We have asked the Code expert to identify potential issues with the function calls in this round and focus on improving the robustness, accuracy, and readability of the code.
%

\subsection{Action Controller}

The code action can leverage these extensive tools to navigate a highly complex action space. However, such complexity may significantly hinder the agent's ability to select the appropriate action. To identify the correct action space and reasoning path, we employ the Action Controller, which determines the most suitable action space and evaluates the reasoning path. The Action Controller consists of two modules: the RAG Selector and the Path Reviewer. The RAG Selector primarily utilizes the Retrieval-Augmented Generation(RAG) \citep{Lewis2020RetrievalAugmentedGF} technique to keep the necessary tools and few-shot examples set accessible to agents as they transition between different reasoning steps. The Path Reviewer manages the reasoning paths for exceptions through three key tasks: handling exceptions, rewriting queries and answers, and reflecting on each reasoning step.

\subsubsection{RAG Selector}

A large number of tools and few-shot examples can expand the action space of the agent, but they may also hinder the agent's decision-making capabilities and increase the coding time of the LLMs, ultimately reducing the agent's performance. To enable the PoAct to execute the most appropriate actions within a complex action space, we employ a RAG Selector to dynamically manage the visible tools and examples, thereby controlling the action space available to the PoAct. For more implementation details you can refer to Appendix \ref{app:implement details}.

\textbf{Tools and Few-Shot Examples} In accordance with various task scenarios, we design different generic tools and few-shot examples for PoAct. LLMs will organize these tools to execute complex tasks based on different reasoning policies, using information about the tools, including their names, descriptions, input cases, and output cases. Few-shot examples primarily consist of instances of error-prone steps alongside exemplary uses of the tools. We generate these few-shot examples based on task types and manually review and modify them as necessary. The RAG Selector will retrieve and rearrange the most relevant tools and few-shot examples, dynamically updating them in the PoAct system prompt throughout the reasoning process. 

\begin{table*}[t]
\renewcommand\arraystretch{1.2}
\resizebox{1.0\linewidth}{!}{
\begin{tabular}{cccccccccc}
\hline
\textbf{llm}                     & \textbf{method}  & \textbf{1-hop (80)} & \textbf{2-hop(80)} & \textbf{3-hop(60)} & \textbf{4-hop(40)} & \textbf{5-hop(20)} & \textbf{knowledge} & \textbf{all}   & \textbf{tokens\_usage} \\ \hline
\multirow{4}{*}{glm-4-plus}      & PoAct            & \textbf{100.00}     & \textbf{96.88}     & 75.83              & 62.92              & 58.33              & \textbf{85.25}     & 85.63          & 4,008,385              \\
                                 & ReAct            & 82.17               & 73.12              & 33.33              & 27.50              & 14.17              & 42.89              & 59.52          & 185,143,296            \\
                                 & P-S   & 75.81               & 62.71              & 40.00              & 23.75              & 27.50              & 76.02              & 58.94          & 53,611,213             \\
                                 & P-E & 77.69               & 63.54              & 35.00              & 26.25              & 30.83              & 78.16              & 59.39          & 53,404,706             \\ \hline
\multirow{4}{*}{glm-4-air}       & PoAct            & 80.21               & 63.96              & 56.67              & 50.83              & 26.67              & 46.78              & 61.45          & 10,399,287             \\
                                 & ReAct            & 71.12               & 45.21              & 33.33              & 29.58              & 17.50              & 53.80              & 49.70          & 93,853,800             \\
                                 & P-S   & 31.25               & 20.62              & 8.33               & 16.67              & 13.33              & 23.86              & 21.64          & 19,827,659             \\
                                 & P-E & 26.88               & 23.75              & 10.00              & 30.83              & 14.17              & 26.36              & 23.91          & 37,648,236             \\ \hline
\multirow{4}{*}{glm-4-flash}     & PoAct            & 74.75               & 73.96              & 40.00              & 29.17              & 31.83              & 46.07              & 56.74          & 13,187,633             \\
                                 & ReAct            & 73.82               & 44.17              & 20.00              & 18.33              & 13.33              & 24.78              & 43.34          & 109,359,559            \\
                                 & P-S   & 56.44               & 35.00              & 13.33              & 12.08              & 11.67              & 33.95              & 33.97          & 21,091,316             \\
                                 & P-E & 56.19               & 37.92              & 11.67              & 18.33              & 15.00              & 33.91              & 35.50          & 38,062,535             \\ \hline
gpt-4o                           & PoAct            & \textbf{100.00}     & 91.67              & \textbf{83.33}     & 73.75              & 66.83              & 82.22              & \textbf{87.55} & 19,611,017             \\ \hline
gpt-4o-mini                      & PoAct            & 96.25               & 83.54              & 62.50              & 38.75              & 27.50              & 83.48              & 73.01          & 21,891,578             \\ \hline
\multicolumn{1}{l}{qwen-2.5-72b} & PoAct            & 97.50               & 92.50              & 71.67              & \textbf{75.83}     & \textbf{74.17}     & 84.45              & 85.69          & -                      \\ \hline
\end{tabular}
}
\caption{The main experimental results on the LegalAgentBench.}
\label{tab:legalagentbench}
\end{table*}

\subsubsection{Path Reviewer}

The complex action space can lead to multiple anomalous reasoning paths, significantly impacting the performance and robustness of the agent's reasoning process. As illustrated in Table \ref{fig:path reviewer}, the Path Reviewer alleviates some of these abnormal reasoning paths through three tasks: \textbf{exception handling}, \textbf{query \& answer rewrite}, and \textbf{code action reflection}.

\textbf{Exception Handling} In the process of solving PoAct tasks from the server side, the execution of code blocks may result in multiple exceptions. To address all potential exceptions, we append a message role error to the exception message and include specific exception details, along with possible causes and solutions, in the Agent's dialog history. These error messages will effectively assist the agent in adopting the correct approach to resolve the current issue, rather than terminating or abandoning the reasoning process upon encountering an exception. By implementing robust exception handling, we can significantly enhance the reliability and fluency of PoAct's problem-solving capabilities.

\textbf{Query \& Answer Rewrite} In order for PoAct to better understand the query, Path Reviewer will extend the query based on its type, such as by incorporating relevant domain knowledge. Query rewriting can effectively assist PoAct in identifying the appropriate reasoning path more quickly. Additionally, to ensure the accuracy and fluency of the agent's responses, the Path Reviewer will review and modify the final responses regarding formatting and clarity. For instance, maintaining the corresponding decimal places as specified in the query, avoiding unusual formatting that may hinder readability, and addressing other potential issues. This review process enhances the quality of the responses and addresses concerns beyond logical errors in reasoning.

\textbf{Code Action Reflection} In addition to rewriting queries and answers, we also aim to reflect and correct PoAct's code actions through reflection rules. These rules primarily include error code backtracking and unanticipated code reflection. Error code backtracking is designed to prevent the agent from repeatedly attempting incorrect code actions. When the agent encounters a mistake that results in the same error consecutively, PoAct will backtrack through the dialogue history to the correct step and adopt a different approach to resolve the issue. For unanticipated code actions, we establish specific triggers to reject unexpected code actions. These triggers can be based on keyword matching or model-based detection. When a trigger identifies code that falls outside of expected parameters, the Path Reviewer will backtrack to the erroneous step and provide a hint message to assist in resolving the issue, guiding PoAct toward the correct reasoning path.

\section{Results and Analysis}
\subsection{Experimental Setup}

\textbf{Datasets} We comprehensively evaluate PoAct's capabilities using two agent datasets. \textbf{1) LegalAgentBench} \citep{author2024repository}: A challenging dataset designed to assess agents' problem-solving abilities in complex legal scenarios. LegalAgentBench is based on real legal cases and encompasses a wide range of intricate tasks, including compliance reviews, contract disputes, and the drafting of legal documents.\textbf{ 2) AgentBench} \citep{Liu2023AgentBenchEL}: This dataset evaluates agents' capabilities across various environments from multiple dimensions. We specifically assessed PoAct's capabilities on \textbf{Database (DB)} and \textbf{Web Browsing (WB)}. For more details about the datasets, please refer to \ref{app:dataset}.

\textbf{Baseline} We use the following agent paradigms as baselines: \textbf{1) ReAct}: single-step execution of the tool until the task is completed, following the thought-action-observation framework \citep{Yao2022ReActSR}. \textbf{2) Plan and Solve (P-S)}: LLMs generate a structured plan based on the user's query, with each step of the plan specifying the tool and the parameters for execution. We execute the tools sequentially according to the plan and allow the LLMs to answer questions based on all observations \citep{Xu2023ReWOODR}. \textbf{3) Plan and Execute (P-E)}: LLMs generate a list of tasks and execute the tool according to each step in a single action, updating the plan based on the results of the execution until the task is completed\citep{Wang2023PlanandSolvePI}. We maintain the same tool settings and descriptions as PoAct and use an equivalent number of few-shot examples.

\textbf{LLMs} We conduct experiments on 6 LLMs, including OpenAI's GPT-4o (gpt-4o-2024-08-06) and GPT-4o-mini (gpt-4o-mini-2024-07-18) \citep{Achiam2023GPT4TR}, three glm-4 family LLMs (glm-4-plus, glm-4-air, glm-4-flash) \citep{glm2024chatglm}, and an open-source model, Qwen-2.5-72b-instruct \citep{qwen2.5}.

\textbf{Evaluation Metric} In LegalAgentBench, we utilize the \textit{Success Rate (SR)} to quantify the precision of the correct keyword in the final answer. We employ SR and Step SR to assess the success rate of PoAct on the WB and DB datasets.

\begin{table*}[t]
\renewcommand\arraystretch{1}
\resizebox{1.0\linewidth}{!}{
\begin{tabular}{ccccccccc}
\hline
\textbf{few shots}  & \textbf{tools} & \textbf{1-hop (80)} & \textbf{2-hop(80)} & \textbf{3-hop(60)} & \textbf{4-hop(40)} & \textbf{5-hop(20)} & \textbf{knowledge} & \textbf{all}   \\ \hline
0                   & 5              & 98.23               & 87.50              & 70.83              & 50.00              & \textbf{62.33}     & 86.37              & 80.23          \\
0                   & 10             & 93.94               & 91.04              & 70.00              & 58.75              & 44.33              & 85.38              & 79.81          \\
1                   & 5              & 95.70               & 90.62              & \textbf{78.33}     & 60.42              & 57.67              & 80.00              & 82.53          \\
1                   & 10             & 94.22               & 94.17              & 74.17              & \textbf{67.50}     & 52.17              & \textbf{86.85}     & 83.31          \\
3                   & 5              & \textbf{100.00}     & \textbf{96.88}     & 75.83              & 62.92              & 58.33              & 85.25              & \textbf{85.63} \\
3                   & 10             & 94.62               & 88.75              & 74.17              & 62.92              & 52.00              & 84.44              & 81.22          \\
\multicolumn{2}{c}{w/o rag selector} & 91.25               & 84.38              & 66.67              & 66.25              & 34.33              & 77.74              & 76.47          \\ \hline
\end{tabular}
}
\caption{The success rate of PoAct on LegalAgentBench with different few-shot examples and tools settings and without the RAG selector.}
\label{tab:rag selector}
\end{table*}

\subsection{Experiments on LegalAgentBench}

\textbf{Main Experimental Results} Table \ref{tab:legalagentbench} presents the primary evaluation results for the gpt-4o, glm-4 family of models, and qwen-72b-instruct on LegalAgentBench. Our observations indicate that PoAct consistently outperforms the baseline scores across all problem types. Furthermore, PoAct consumes far fewer tokens than the baseline approach in numerous tools setting. These results underscore the effectiveness of PoAct in enhancing the capabilities of agents in complex scenarios, thereby improving the performance of agent systems. In the glm-4-plus setting, PoAct's success rate is 25 percent higher than the ReAct, with the advantage of PoAct becoming increasingly pronounced as problem difficulty escalates. Notably, the success rate of PoAct in 5-hop problems is 28 percent superior to that of P-E, demonstrating the potential of our method in addressing multi-hop complex reasoning problems.

\subsubsection{Ablation Study}
To evaluate the capabilities of the modules, we conducted ablation experiments on glm-4-plus. As illustrated in Table \ref{tab:rag selector}, we compare the results of experiments under various settings of few-shot examples and tools to assess the impact of the RAG selector. Our observations indicate that a limited amount of context enhances the agent's reasoning ability. Conversely, when the context is excessively lengthy, an overload of information regarding the few-shot examples and tools may hinder the agent's reasoning performance.

\begin{table}[h]
\renewcommand\arraystretch{1}
\resizebox{1.0\linewidth}{!}{
\begin{tabular}{cccccccc}
\hline
\textbf{method}             & \textbf{1-hop (80)} & \textbf{2-hop(80)} & \textbf{3-hop(60)} & \textbf{4-hop(40)} & \textbf{5-hop(20)} & \textbf{knowledge} & \textbf{all}   \\ \hline
PoAct                       & \textbf{100.00}     & \textbf{96.88}     & 75.83              & \textbf{62.92}     & \textbf{58.33}     & 85.25              & \textbf{85.63} \\
w/o QAR & 86.88               & 80.62              & \textbf{80.00}     & 62.50              & 54.17              & \textbf{85.77}     & 78.33          \\
w/o CAR & 95.31               & 89.38              & 74.17              & 60.83              & 50.00              & 75.78              & 80.58          \\
w/o QAR \& CAR           & 79.12               & 82.50              & 78.33              & 51.67              & 41.00              & 79.58              & 73.69          \\ \hline
\end{tabular}
}
\caption{The ablation studies on the path reviewer.}
\label{tab:path reviewer}
\end{table}

As illustrated in Table \ref{tab:path reviewer}, we conducted ablation experiments on the Code Action Reflections (CAR) and Query \& Answer Rewrite (QAR) tasks of the path reviewer. The results indicate that Query \& Answer Rewrite can effectively mitigate the formatting and fluency issues present in both the query and answer. However, in the case of 3-hop and knowledge-based problems, we observed that Query \& Answer Rewrite may result in the loss of some semantic information during the rewriting leading to poorer experimental outcomes. Additionally, we found that Code Action Reflection can effectively prevent the model from falling into anomalous reasoning paths, thereby improving the success rate on complex problems.

\subsection{Experiments on AgentBench}
Table \ref{tab:agentbench} presents the experimental results of PoAct and ReAct for the glm-4 family of models in web browsing and database environments. Our findings indicate that PoAct consistently outperforms ReAct across all environments, achieving a 12.2 percent improvement on the mind2web task in the glm-4-plus setup. This suggests that the enhancement in PoAct's success rate is reliable across various environments.

\begin{table}[h]
\renewcommand\arraystretch{1}
\resizebox{1.0\linewidth}{!}{
\begin{tabular}{cccc}
\hline
\textbf{llm}                & \textbf{method} & \textbf{database} & \textbf{mind2web} \\ \hline
\multirow{2}{*}{glm-4-plus} & PoAct         & \textbf{67.3 }             & \textbf{36.3}              \\
                            & ReAct         & 56.0              & 24.0              \\ \hline
\multirow{2}{*}{glm-4-air}  & PoAct         & 57.2              & 35.4              \\
                            & ReAct         & 51                & 30.0              \\ \hline
\end{tabular}
}
\caption{Experimental results of PoAct and ReAct in web browsing and database environments.}
\label{tab:agentbench}
\end{table}

\subsection{Case Study}

We found that PoAct exhibits strong generalization capabilities. With a simple tool description customized to the data source, PoAct can solve most complex tasks with zero or few shots. In our experiments, we discovered that, thanks to the RAG Selector, the addition of few-shot examples and tools has little impact on the performance and accuracy of PoAct, demonstrating excellent scalability in specific domains. In practice, we have developed several versions of PoAct for various applications, including general assistance and data analysis. In real business scenarios, PoAct demonstrates remarkable scalability and versatility. More examples can be found in Appendix \ref{app:case trajectories}.

\section{Conclusion}

We propose PoAct, a code agent that employs dual control over policy and action to execute high-quality code actions and optimize inference paths by dynamically switching between inference policies and action spaces. We evaluate our approach using two challenging agent benchmarks. The results indicate that PoAct significantly enhances the reasoning capabilities of tool-calling agents when addressing complex tasks, while also substantially reducing token consumption. Furthermore, we present various case studies of PoAct across different scenarios, demonstrating its strong scalability and generalizability.

\section*{Limitations}

This study has two limitations. First, regarding the RAG Selector, it does not compare more complex and diverse embedded models to assess the impact of the RAG system on the experimental results. Second, we observed that the Code Agent appears to lose the language models' capability for complex reasoning. For instance, the Code Agent often resorts to using code to filter specific fields from structured data, even when the value of the field has already been presented in the context.

\bibliography{custom}

\begin{thebibliography}{29}
\providecommand{\natexlab}[1]{#1}

\bibitem[{Achiam et~al.(2023)Achiam, Adler et~al.}]{Achiam2023GPT4TR}
OpenAI~Josh Achiam, Steven Adler, et~al. 2023.
\newblock \href {https://api.semanticscholar.org/CorpusID:257532815} {Gpt-4 technical report}.

\bibitem[{Floridi and Chiriatti(2020)}]{Floridi2020GPT3IN}
L.~Floridi and Massimo Chiriatti. 2020.
\newblock \href {https://api.semanticscholar.org/CorpusID:228954221} {Gpt-3: Its nature, scope, limits, and consequences}.
\newblock \emph{Minds and Machines}, 30:681 -- 694.

\bibitem[{Gao et~al.(2023)Gao, Ji, Zhou, Lin, Chen, Fan, and Shou}]{Gao2023AssistGPTAG}
Difei Gao, Lei Ji, Luowei Zhou, Kevin Lin, Joya Chen, Zihan Fan, and Mike~Zheng Shou. 2023.
\newblock \href {https://api.semanticscholar.org/CorpusID:259164559} {Assistgpt: A general multi-modal assistant that can plan, execute, inspect, and learn}.
\newblock \emph{ArXiv}, abs/2306.08640.

\bibitem[{GLM et~al.(2024)GLM, Zeng, Xu, Wang, Zhang, Yin, Rojas, Feng, Zhao, Lai, Yu, Wang, Sun, Zhang, Cheng, Gui, Tang, Zhang, Li, Zhao, Wu, Zhong, Liu, Huang, Zhang, Zheng, Lu, Duan, Zhang, Cao, Yang, Tam, Zhao, Liu, Xia, Zhang, Gu, Lv, Liu, Liu, Yang, Song, Zhang, An, Xu, Niu, Yang, Li, Bai, Dong, Qi, Wang, Yang, Du, Hou, and Wang}]{glm2024chatglm}
Team GLM, Aohan Zeng, Bin Xu, Bowen Wang, Chenhui Zhang, Da~Yin, Diego Rojas, Guanyu Feng, Hanlin Zhao, Hanyu Lai, Hao Yu, Hongning Wang, Jiadai Sun, Jiajie Zhang, Jiale Cheng, Jiayi Gui, Jie Tang, Jing Zhang, Juanzi Li, Lei Zhao, Lindong Wu, Lucen Zhong, Mingdao Liu, Minlie Huang, Peng Zhang, Qinkai Zheng, Rui Lu, Shuaiqi Duan, Shudan Zhang, Shulin Cao, Shuxun Yang, Weng~Lam Tam, Wenyi Zhao, Xiao Liu, Xiao Xia, Xiaohan Zhang, Xiaotao Gu, Xin Lv, Xinghan Liu, Xinyi Liu, Xinyue Yang, Xixuan Song, Xunkai Zhang, Yifan An, Yifan Xu, Yilin Niu, Yuantao Yang, Yueyan Li, Yushi Bai, Yuxiao Dong, Zehan Qi, Zhaoyu Wang, Zhen Yang, Zhengxiao Du, Zhenyu Hou, and Zihan Wang. 2024.
\newblock \href {https://arxiv.org/abs/2406.12793} {Chatglm: A family of large language models from glm-130b to glm-4 all tools}.
\newblock \emph{Preprint}, arXiv:2406.12793.

\bibitem[{Johnson et~al.(2019)Johnson, Douze, and J{\'e}gou}]{johnson2019billion}
Jeff Johnson, Matthijs Douze, and Herv{\'e} J{\'e}gou. 2019.
\newblock Billion-scale similarity search with {GPUs}.
\newblock \emph{IEEE Transactions on Big Data}, 7(3):535--547.

\bibitem[{Lewis et~al.(2020)Lewis, Perez, Piktus, Petroni, Karpukhin, Goyal, Kuttler, Lewis, tau Yih, Rockt{\"a}schel, Riedel, and Kiela}]{Lewis2020RetrievalAugmentedGF}
Patrick Lewis, Ethan Perez, Aleksandara Piktus, Fabio Petroni, Vladimir Karpukhin, Naman Goyal, Heinrich Kuttler, Mike Lewis, Wen tau Yih, Tim Rockt{\"a}schel, Sebastian Riedel, and Douwe Kiela. 2020.
\newblock \href {https://api.semanticscholar.org/CorpusID:218869575} {Retrieval-augmented generation for knowledge-intensive nlp tasks}.
\newblock \emph{ArXiv}, abs/2005.11401.

\bibitem[{Li et~al.(2024)Li, Chen, Yang, Ai, Jia, youfeng Liu, Lin, WU, Yuan, HU, Wang, LIU, and Huang}]{author2024repository}
Haitao Li, Junjie Chen, Jingli Yang, Qingyao Ai, Wei Jia, youfeng Liu, Kai Lin, Yueyue WU, Guozhi Yuan, Yiran HU, Wuyue Wang, Yiqun LIU, and Minlie Huang. 2024.
\newblock \href {https://github.com/CSHaitao/LegalAgentBench} {Legalagentbench: Evaluating llm agents in legal domain}.
\newblock Accessed: 2024-12-13.

\bibitem[{Liu et~al.(2023)Liu, Yu, Zhang, Xu, Lei, Lai, Gu, Gu, Ding, Men, Yang, Zhang, Deng, Zeng, Du, Zhang, Shen, Zhang, Shen, Su, Sun, Huang, Dong, and Tang}]{Liu2023AgentBenchEL}
Xiao Liu, Hao Yu, Hanchen Zhang, Yifan Xu, Xuanyu Lei, Hanyu Lai, Yu~Gu, Yuxian Gu, Hangliang Ding, Kai Men, Kejuan Yang, Shudan Zhang, Xiang Deng, Aohan Zeng, Zhengxiao Du, Chenhui Zhang, Shengqi Shen, Tianjun Zhang, Sheng Shen, Yu~Su, Huan Sun, Minlie Huang, Yuxiao Dong, and Jie Tang. 2023.
\newblock \href {https://api.semanticscholar.org/CorpusID:260682249} {Agentbench: Evaluating llms as agents}.
\newblock \emph{ArXiv}, abs/2308.03688.

\bibitem[{Ouyang et~al.(2022)Ouyang, Wu, Jiang, Almeida, Wainwright, Mishkin, Zhang, Agarwal, Slama, Ray, Schulman, Hilton, Kelton, Miller, Simens, Askell, Welinder, Christiano, Leike, and Lowe}]{Ouyang2022TrainingLM}
Long Ouyang, Jeff Wu, Xu~Jiang, Diogo Almeida, Carroll~L. Wainwright, Pamela Mishkin, Chong Zhang, Sandhini Agarwal, Katarina Slama, Alex Ray, John Schulman, Jacob Hilton, Fraser Kelton, Luke~E. Miller, Maddie Simens, Amanda Askell, Peter Welinder, Paul~Francis Christiano, Jan Leike, and Ryan~J. Lowe. 2022.
\newblock \href {https://api.semanticscholar.org/CorpusID:246426909} {Training language models to follow instructions with human feedback}.
\newblock \emph{ArXiv}, abs/2203.02155.

\bibitem[{Park et~al.(2023)Park, O'Brien, Cai, Morris, Liang, and Bernstein}]{Park2023GenerativeAI}
Joon~Sung Park, Joseph~C. O'Brien, Carrie~J. Cai, Meredith~Ringel Morris, Percy Liang, and Michael~S. Bernstein. 2023.
\newblock \href {https://api.semanticscholar.org/CorpusID:258040990} {Generative agents: Interactive simulacra of human behavior}.
\newblock \emph{Proceedings of the 36th Annual ACM Symposium on User Interface Software and Technology}.

\bibitem[{Qiao et~al.(2023)Qiao, Li, Zhang, He, Kang, Zhang, Yang, Dong, Zhang, Wang, Ma, Zhao, Qin, Qin, Du, Xu, Lin, Rajmohan, and Zhang}]{Qiao2023TaskWeaverAC}
Bo~Qiao, Liqun Li, Xu~Zhang, Shilin He, Yu~Kang, Chaoyun Zhang, Fangkai Yang, Hang Dong, Jue Zhang, Lu~Wang, Ming-Jie Ma, Pu~Zhao, Si~Qin, Xiaoting Qin, Chao Du, Yong Xu, Qingwei Lin, S.~Rajmohan, and Dongmei Zhang. 2023.
\newblock \href {https://api.semanticscholar.org/CorpusID:265498341} {Taskweaver: A code-first agent framework}.
\newblock \emph{ArXiv}, abs/2311.17541.

\bibitem[{Qin et~al.(2023)Qin, Liang, Ye, Zhu, Yan, Lu, Lin, Cong, Tang, Qian, Zhao, Tian, Xie, Zhou, Gerstein, Li, Liu, and Sun}]{Qin2023ToolLLMFL}
Yujia Qin, Shi Liang, Yining Ye, Kunlun Zhu, Lan Yan, Ya-Ting Lu, Yankai Lin, Xin Cong, Xiangru Tang, Bill Qian, Sihan Zhao, Runchu Tian, Ruobing Xie, Jie Zhou, Marc~H. Gerstein, Dahai Li, Zhiyuan Liu, and Maosong Sun. 2023.
\newblock \href {https://api.semanticscholar.org/CorpusID:260334759} {Toolllm: Facilitating large language models to master 16000+ real-world apis}.
\newblock \emph{ArXiv}, abs/2307.16789.

\bibitem[{{Qwen Team}(2024)}]{qwen2.5}
{Qwen Team}. 2024.
\newblock \href {https://qwenlm.github.io/blog/qwen2.5/} {Qwen2.5: A party of foundation models}.

\bibitem[{Radford et~al.(2019)Radford, Wu, Child, Luan, Amodei, and Sutskever}]{Radford2019LanguageMA}
Alec Radford, Jeff Wu, Rewon Child, David Luan, Dario Amodei, and Ilya Sutskever. 2019.
\newblock \href {https://api.semanticscholar.org/CorpusID:160025533} {Language models are unsupervised multitask learners}.

\bibitem[{Thoppilan et~al.(2022)Thoppilan, Freitas, Hall, Shazeer, Kulshreshtha, Cheng, Jin, Bos, Baker, Du, Li, Lee, Zheng, Ghafouri, Menegali, Huang, Krikun, Lepikhin, Qin, Chen, Xu, Chen, Roberts, Bosma, Zhou, Chang, Krivokon, Rusch, Pickett, Meier-Hellstern, Morris, Doshi, Santos, Duke, S{\o}raker, Zevenbergen, Prabhakaran, D{\'i}az, Hutchinson, Olson, Molina, Hoffman-John, Lee, Aroyo, Rajakumar, Butryna, Lamm, Kuzmina, Fenton, Cohen, Bernstein, Kurzweil, Aguera-Arcas, Cui, Croak, Chi, and Le}]{Thoppilan2022LaMDALM}
Romal Thoppilan, Daniel~De Freitas, Jamie Hall, Noam~M. Shazeer, Apoorv Kulshreshtha, Heng-Tze Cheng, Alicia Jin, Taylor Bos, Leslie Baker, Yu~Du, Yaguang Li, Hongrae Lee, Huaixiu~Steven Zheng, Amin Ghafouri, Marcelo Menegali, Yanping Huang, Maxim Krikun, Dmitry Lepikhin, James Qin, Dehao Chen, Yuanzhong Xu, Zhifeng Chen, Adam Roberts, Maarten Bosma, Yanqi Zhou, Chung-Ching Chang, I.~A. Krivokon, Willard~James Rusch, Marc Pickett, Kathleen~S. Meier-Hellstern, Meredith~Ringel Morris, Tulsee Doshi, Renelito~Delos Santos, Toju Duke, Johnny~Hartz S{\o}raker, Ben Zevenbergen, Vinodkumar Prabhakaran, Mark D{\'i}az, Ben Hutchinson, Kristen Olson, Alejandra Molina, Erin Hoffman-John, Josh Lee, Lora Aroyo, Ravi Rajakumar, Alena Butryna, Matthew Lamm, V.~O. Kuzmina, Joseph Fenton, Aaron Cohen, Rachel Bernstein, Ray Kurzweil, Blaise Aguera-Arcas, Claire Cui, Marian~Rogers Croak, Ed~H. Chi, and Quoc Le. 2022.
\newblock \href {https://api.semanticscholar.org/CorpusID:246063428} {Lamda: Language models for dialog applications}.
\newblock \emph{ArXiv}, abs/2201.08239.

\bibitem[{Wang et~al.(2023{\natexlab{a}})Wang, Xie, Jiang, Mandlekar, Xiao, Zhu, Fan, and Anandkumar}]{Wang2023VoyagerAO}
Guanzhi Wang, Yuqi Xie, Yunfan Jiang, Ajay Mandlekar, Chaowei Xiao, Yuke Zhu, Linxi~(Jim) Fan, and Anima Anandkumar. 2023{\natexlab{a}}.
\newblock \href {https://api.semanticscholar.org/CorpusID:258887849} {Voyager: An open-ended embodied agent with large language models}.
\newblock \emph{Trans. Mach. Learn. Res.}, 2024.

\bibitem[{Wang et~al.(2023{\natexlab{b}})Wang, Xu, Lan, Hu, Lan, Lee, and Lim}]{Wang2023PlanandSolvePI}
Lei Wang, Wanyu Xu, Yihuai Lan, Zhiqiang Hu, Yunshi Lan, Roy Ka-Wei Lee, and Ee-Peng Lim. 2023{\natexlab{b}}.
\newblock \href {https://api.semanticscholar.org/CorpusID:258558102} {Plan-and-solve prompting: Improving zero-shot chain-of-thought reasoning by large language models}.
\newblock In \emph{Annual Meeting of the Association for Computational Linguistics}.

\bibitem[{Wang et~al.(2024)Wang, Wang, Su, Tong, and Song}]{Wang2024RethinkingTB}
Qineng Wang, Zihao Wang, Ying Su, Hanghang Tong, and Yangqiu Song. 2024.
\newblock \href {https://api.semanticscholar.org/CorpusID:268041461} {Rethinking the bounds of llm reasoning: Are multi-agent discussions the key?}
\newblock In \emph{Annual Meeting of the Association for Computational Linguistics}.

\bibitem[{Wang et~al.(2022)Wang, Li, and Ji}]{Wang2022Code4StructCG}
Xingyao Wang, Sha Li, and Heng Ji. 2022.
\newblock \href {https://api.semanticscholar.org/CorpusID:258887711} {Code4struct: Code generation for few-shot event structure prediction}.
\newblock In \emph{Annual Meeting of the Association for Computational Linguistics}.

\bibitem[{Wei et~al.(2021)Wei, Bosma, Zhao, Guu, Yu, Lester, Du, Dai, and Le}]{Wei2021FinetunedLM}
Jason Wei, Maarten Bosma, Vincent Zhao, Kelvin Guu, Adams~Wei Yu, Brian Lester, Nan Du, Andrew~M. Dai, and Quoc~V. Le. 2021.
\newblock \href {https://api.semanticscholar.org/CorpusID:237416585} {Finetuned language models are zero-shot learners}.
\newblock \emph{ArXiv}, abs/2109.01652.

\bibitem[{Wei et~al.(2022)Wei, Wang, Schuurmans, Bosma, Chi, Xia, Le, and Zhou}]{Wei2022ChainOT}
Jason Wei, Xuezhi Wang, Dale Schuurmans, Maarten Bosma, Ed~H. Chi, F.~Xia, Quoc Le, and Denny Zhou. 2022.
\newblock \href {https://api.semanticscholar.org/CorpusID:246411621} {Chain of thought prompting elicits reasoning in large language models}.
\newblock \emph{ArXiv}, abs/2201.11903.

\bibitem[{Wolf et~al.(2020)Wolf, Debut, Sanh, Chaumond, Delangue, Moi, Cistac, Rault, Louf, Funtowicz, Davison, Shleifer, von Platen, Ma, Jernite, Plu, Xu, Scao, Gugger, Drame, Lhoest, and Rush}]{wolf-etal-2020-transformers}
Thomas Wolf, Lysandre Debut, Victor Sanh, Julien Chaumond, Clement Delangue, Anthony Moi, Pierric Cistac, Tim Rault, Rémi Louf, Morgan Funtowicz, Joe Davison, Sam Shleifer, Patrick von Platen, Clara Ma, Yacine Jernite, Julien Plu, Canwen Xu, Teven~Le Scao, Sylvain Gugger, Mariama Drame, Quentin Lhoest, and Alexander~M. Rush. 2020.
\newblock \href {https://www.aclweb.org/anthology/2020.emnlp-demos.6} {Transformers: State-of-the-art natural language processing}.
\newblock In \emph{Proceedings of the 2020 Conference on Empirical Methods in Natural Language Processing: System Demonstrations}, pages 38--45, Online. Association for Computational Linguistics.

\bibitem[{Xi et~al.(2024)Xi, Ding, Chen, Hong, Guo, Wang, Yang, Liao, Guo, He, Gao, Chen, Zheng, Zou, Gui, Zhang, Qiu, Huang, Wu, and Jiang}]{Xi2024AgentGymEL}
Zhiheng Xi, Yiwen Ding, Wenxiang Chen, Boyang Hong, Honglin Guo, Junzhe Wang, Dingwen Yang, Chenyang Liao, Xin Guo, Wei He, Songyang Gao, Luyao Chen, Rui Zheng, Yicheng Zou, Tao Gui, Qi~Zhang, Xipeng Qiu, Xuanjing Huang, Zuxuan Wu, and Yu-Gang Jiang. 2024.
\newblock \href {https://api.semanticscholar.org/CorpusID:270285866} {Agentgym: Evolving large language model-based agents across diverse environments}.
\newblock \emph{ArXiv}, abs/2406.04151.

\bibitem[{Xiao et~al.(2023)Xiao, Liu, Zhang, and Muennighoff}]{bge_embedding}
Shitao Xiao, Zheng Liu, Peitian Zhang, and Niklas Muennighoff. 2023.
\newblock \href {https://arxiv.org/abs/2309.07597} {C-pack: Packaged resources to advance general chinese embedding}.
\newblock \emph{Preprint}, arXiv:2309.07597.

\bibitem[{Xu et~al.(2023)Xu, Peng, Lei, Mukherjee, Liu, and Xu}]{Xu2023ReWOODR}
Binfeng Xu, Zhiyuan Peng, Bowen Lei, Subhabrata Mukherjee, Yuchen Liu, and Dongkuan Xu. 2023.
\newblock \href {https://api.semanticscholar.org/CorpusID:258967566} {Rewoo: Decoupling reasoning from observations for efficient augmented language models}.
\newblock \emph{ArXiv}, abs/2305.18323.

\bibitem[{Yao et~al.(2022)Yao, Zhao, Yu, Du, Shafran, Narasimhan, and Cao}]{Yao2022ReActSR}
Shunyu Yao, Jeffrey Zhao, Dian Yu, Nan Du, Izhak Shafran, Karthik Narasimhan, and Yuan Cao. 2022.
\newblock \href {https://api.semanticscholar.org/CorpusID:252762395} {React: Synergizing reasoning and acting in language models}.
\newblock \emph{ArXiv}, abs/2210.03629.

\bibitem[{Zhang et~al.(2024)Zhang, Li, Li, Shi, and Jin}]{Zhang2024CodeAgentEC}
Kechi Zhang, Jia Li, Ge~Li, Xianjie Shi, and Zhi Jin. 2024.
\newblock \href {https://api.semanticscholar.org/CorpusID:266999556} {Codeagent: Enhancing code generation with tool-integrated agent systems for real-world repo-level coding challenges}.
\newblock In \emph{Annual Meeting of the Association for Computational Linguistics}.

\bibitem[{Zhao et~al.(2023)Zhao, Zhou, Li, Tang, Wang, Hou, Min, Zhang, Zhang, Dong, Du, Yang, Chen, Chen, Jiang, Ren, Li, Tang, Liu, Liu, Nie, and rong Wen}]{Zhao2023ASO}
Wayne~Xin Zhao, Kun Zhou, Junyi Li, Tianyi Tang, Xiaolei Wang, Yupeng Hou, Yingqian Min, Beichen Zhang, Junjie Zhang, Zican Dong, Yifan Du, Chen Yang, Yushuo Chen, Z.~Chen, Jinhao Jiang, Ruiyang Ren, Yifan Li, Xinyu Tang, Zikang Liu, Peiyu Liu, Jianyun Nie, and Ji~rong Wen. 2023.
\newblock \href {https://api.semanticscholar.org/CorpusID:257900969} {A survey of large language models}.
\newblock \emph{ArXiv}, abs/2303.18223.

\bibitem[{Zhu et~al.(2023)Zhu, Chen, Tian, Tao, Su, Yang, Huang, Li, Lu, Wang, Qiao, Zhang, and Dai}]{Zhu2023GhostIT}
Xizhou Zhu, Yuntao Chen, Hao Tian, Chenxin Tao, Weijie Su, Chenyu Yang, Gao Huang, Bin Li, Lewei Lu, Xiaogang Wang, Y.~Qiao, Zhaoxiang Zhang, and Jifeng Dai. 2023.
\newblock \href {https://api.semanticscholar.org/CorpusID:258959262} {Ghost in the minecraft: Generally capable agents for open-world environments via large language models with text-based knowledge and memory}.
\newblock \emph{ArXiv}, abs/2305.17144.

\end{thebibliography}
\appendix

\section{Implementation Details}
\label{app:implement details}

\textbf{Retrieval-Augmented Generation} We use FAISS\citep{johnson2019billion} as the engine for vector retrieval. Algorithmically, we use bge-large-zh-v1.5 as the embedding model for Chinese data and bge-reranker-large as the rearrangement model to accomplish the recall of relevant tools and few-shot examples \citep{bge_embedding}. According to different task scenarios, we maintain a tools database and a few-shot examples database.

\textbf{LLMs Setting} We set temperature to 0 and use the default parameters for everything else.

\section{Step Policy Prompts}

We show the step policy prompts for Planning \ref{tab:planning prompt}, Thought \ref{tab:thought prompt} and Code \ref{tab:code prompt}.

\begin{table*}[h]
\renewcommand\arraystretch{1.1}
\resizebox{1.0\linewidth}{!}{
\large
\begin{tabular}{l}
\hline
\textbf{Planning Policy Prompt}\\ 
\hline
You are an expert in the use of Tools and are adept at using them to formulate efficient Plans to solve a task.\\
\\
\# What you know\\
<<agent\_policy>>\\
\\
\# What you need to do\\
- Please generate a list of Plans based on the task, the Plan contains multiple steps, each of which needs to be completed using a tool.\\
- The Plan must be based on existing Tools to complete the task, which if executed correctly will result in the correct answer.\\
- Each step of the Plan must specify which Tool to use to do, for example, ‘you can use get\_company\_info to get company information’.\\
- Plan only need to specify what Tool is needed and what to solve, do not show the details of the use of Tool and code.\\
- Require complete semantics for each step, don't omit semantic information.\\
- Don't skip any steps and don't add any extra steps.\\
- Output only the result, don't do any explanation!\\
\\
You can use the following tools:\\
<<tool\_descriptions>>\\

You can refer to the following examples to solve the problem, and you must follow the same format: \\
<<few\_shots>> \\
\hline
\end{tabular}
}
\caption{}
\label{tab:planning prompt}
\end{table*}

\begin{table*}[h]
\renewcommand\arraystretch{1.1}
\resizebox{1.0\linewidth}{!}{
\large
\begin{tabular}{l}
\hline
\textbf{Thought Policy Prompt}\\ 
\hline
You are an expert at solving problems and can break down any task into manageable\\ steps through careful analysis. You will be given a task to complete to the best of your ability. \\
\\
\# What you know\\
<<agent\_policy>>\\
\\
\# What you need to do\\
- You can use a set of tools, which are Python functions that can be called with code.\\
- You need to generate a sequence of <thought></thought> elements, in which you need to explain the reason \\
for performing this step and specify the tool you want to use to guide the subsequent <code></code> generation \\
to produce the correct code.\\
- Finally, you must use the 'final\_answer' tool to return the final answer. \\
\\
\# Things to Note\\
- Always provide a <thought></thought> sequence that includes local planning for the current round, with the \\
names of the tools you plan to use.\\
- Be careful not to use multiple Tools in the same thought. Especially when the code output format is unpredictable. \\
 Instead, observe the results of <observation></observation> step by step and provide\\
a new thought based on the results.\\
- You can use Python packages in your thought, but only from the following list of authorized imports: <<authorized\_imports>>.\\
- Don't give up! You are responsible for solving the task, not just providing a direction to solve it.\\
- Use the results of the previous <observation></observation> step to support your answer, don't make up thoughts.\\
- If the <observation></observation> result is empty, you can try using another tool to complete the task. \\

You can use the following tools:\\
<<tool\_descriptions>>\\

You can refer to the following examples to solve the problem, and you must follow the same format: \\
<<few\_shots>> \\
\hline
\end{tabular}
}
\caption{}
\label{tab:thought prompt}
\end{table*}

\begin{table*}[h]
\renewcommand\arraystretch{1.1}
\resizebox{1.0\linewidth}{!}{
\begin{tabular}{l}
\hline
\textbf{Code Policy Prompt}\\ 
\hline
You are a Code Expert  and can solve any task using code blocks. You will generate high-quality Python code for the user's task in each round \\ 
and use the print() function to display the execution results for the user. \\
\\
\# What you know \\
<<agent\_policy>>\\
\\
\# What you need to do\\
- Generate high-quality code using the given Python functions and the user's current task, and display the execution result using the print() function.\\
- When writing the code, you need to consider both the final goal <final\_goal> and the current task <task>. \\
\\
\# Things to Note\\
- Always provide a <code></code> sequence.\\
- Only use variables and Python functions defined by yourself!\\
- Avoid passing parameters in the form of dict like 'answer = ask\_search\_agent({'query': 'James Bond lives where?'}) '.\\
- Be careful not to call multiple functions in succession in the same code block, especially when the output format is unpredictable. \\
Instead, use print() to output the results, so they can be used in the next <code></code>.\\
- You must use print() to print the execution results at the end of each <code></code>, before the results are displayed in the <observation></observation>.\\
- Results that can be observed directly in the observation can be introduced directly, such as a specific field in the judgment result.\\
- Only call the function when needed, and do not call the same function again with the same parameters.\\
- Do not name any new variable the same as the function name, for example, do not name a variable "final\_answer".\\
- You can use import in the code, but only from the following list of modules: <authorized\_imports>.\\
- The final\_answer function will be used to directly return the final result for the user's final goal.\\
- If the query result is empty, you can try calling other functions to complete the task. \\
\\
\# Rules when writing code  \\
\\
\#\# Code Accuracy  \\
...\\
\\
\#\# Use of preset functions  \\
...\\
\\
\#\# Variable naming \\
...\\
\\
\#\# Code Structuring and Readability  \\
...\\

\#\# Choose the right way to write the code depending on the situation\\
...\\

\#\# When answering a user's question, try to answer using the keywords in the 
...\\

\\
\#\# Output priorities\\
...\\
\\
You can use the following Python functions:\\
<<tool\_descriptions>>\\
\\
You can refer to the following examples to solve the problem, and you must follow the same format: \\
<<few\_shots>>\\
\hline
\end{tabular}
}
\caption{}
\label{tab:code prompt}
\end{table*}

\section{Case Trajectories}
\label{app:case trajectories}

In this section, we show real-life examples of PoAct in different business scenarios. In the case track of generic scenarios, we focus on demonstrating PoAct's capabilities in web search, image understanding and generation, and file processing. In the specialized scenarios, we mainly show PoAct's ability to solve structured data and specialized capabilities.

\subsection{General Cases}
\begin{figure*}[ht]
  \includegraphics[width=1\textwidth]{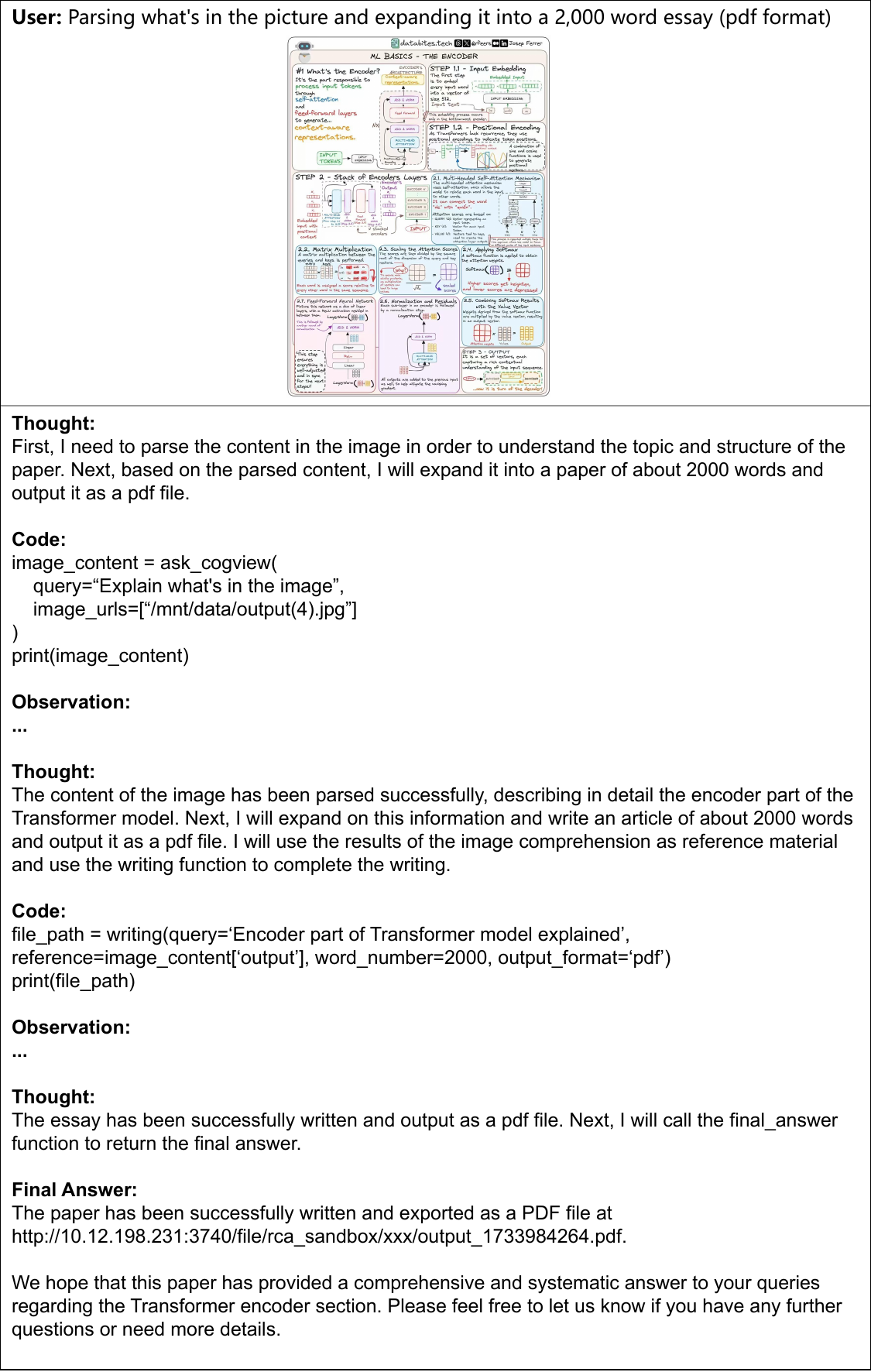}  
  \caption{}
  \label{fig:case1}
\end{figure*}

\begin{figure*}[ht]
  \includegraphics[width=1\textwidth]{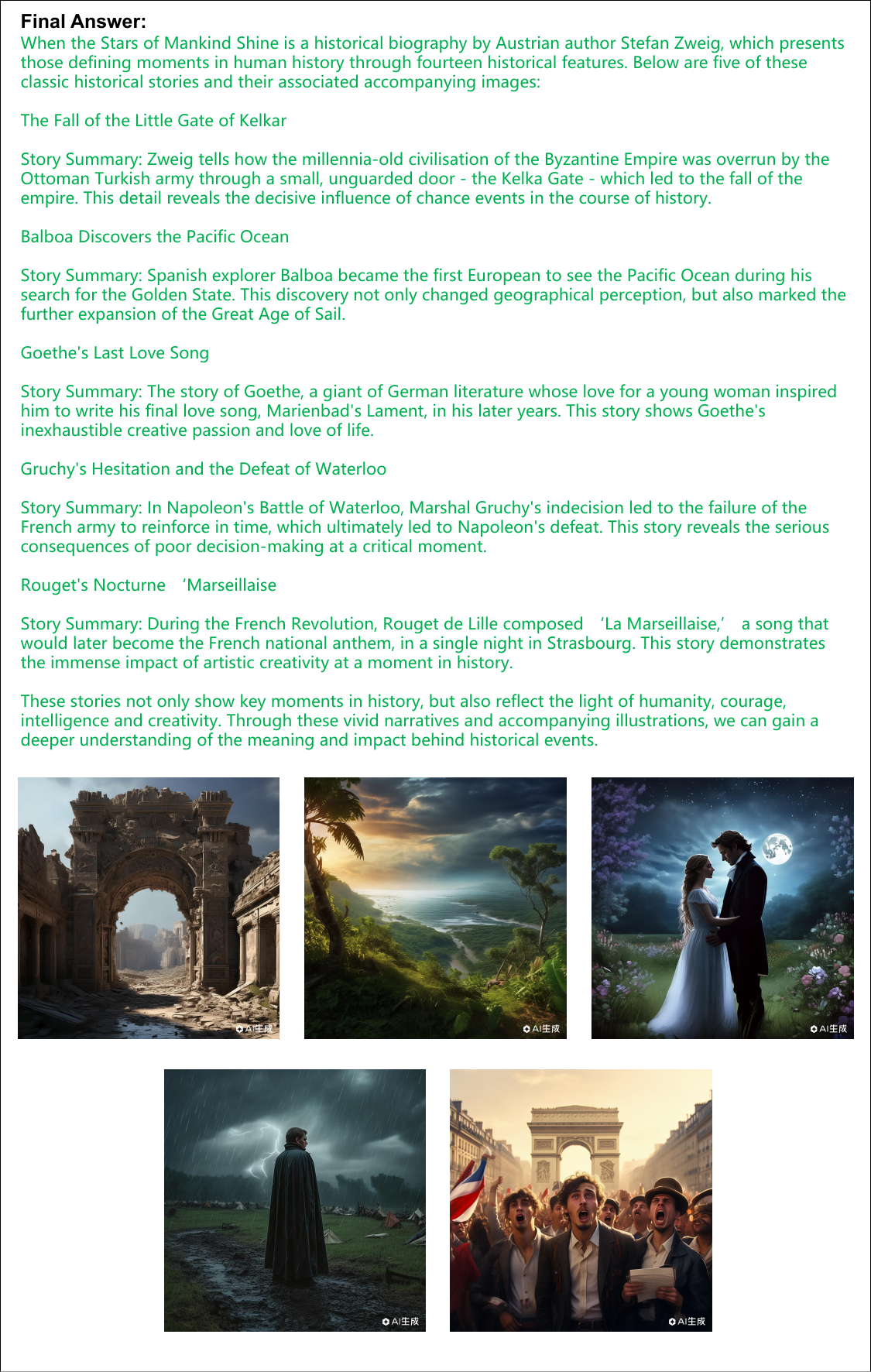} 
  \caption{}
  \label{fig:case2}
\end{figure*}

As shown in Figure \ref{fig:case1}, we asked PoAct to summarise the content of a complex image and output it as a PDF file. As shown in Figure \ref{fig:case2}, we asked PoAct to obtain the content of the book through a web search and generate multiple accompanying images based on the stories searched.

\begin{figure*}[ht]
  \includegraphics[width=1\textwidth]{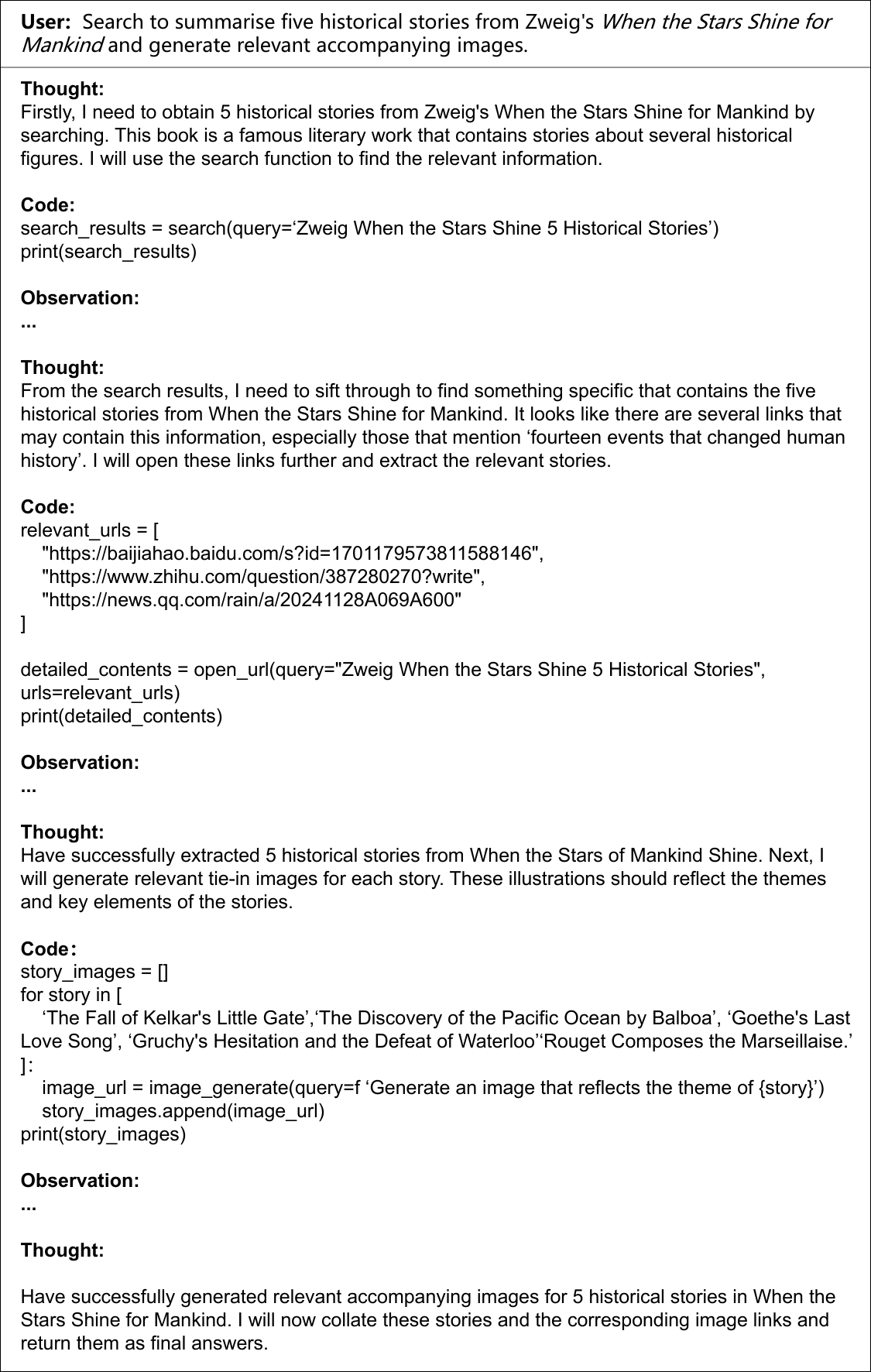} 
\end{figure*}

\subsection{Specialized Cases}
\begin{figure*}[ht]
  \includegraphics[width=1\textwidth]{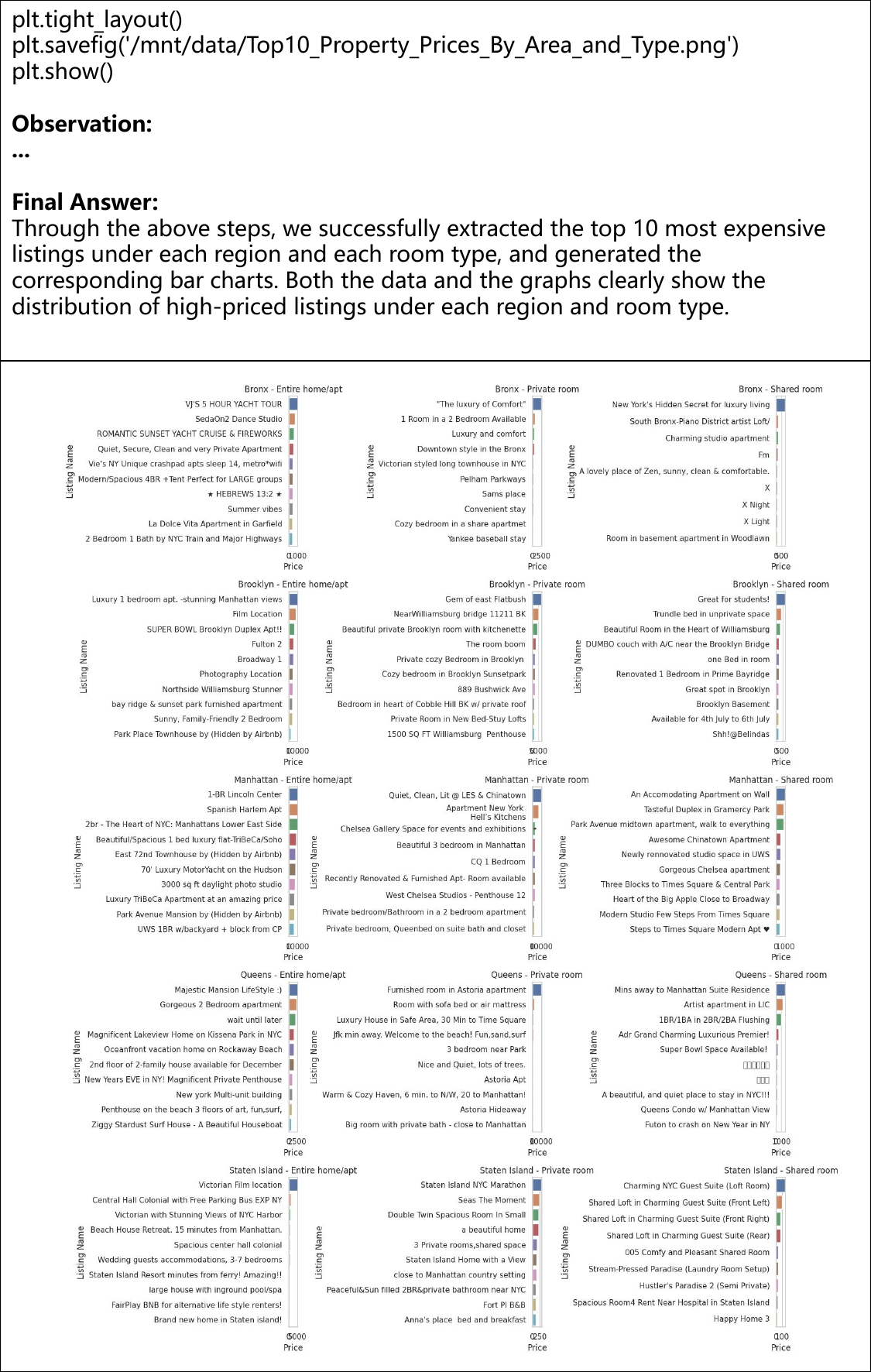} 
  \caption{}
  \label{fig:case3}
\end{figure*}

As shown in Figure \ref{fig:case3}, we asked PoAct to read the table file and analyse the table data as per the requirement and finally present an accurate table.

\begin{figure*}[ht]
  \includegraphics[width=1\textwidth]{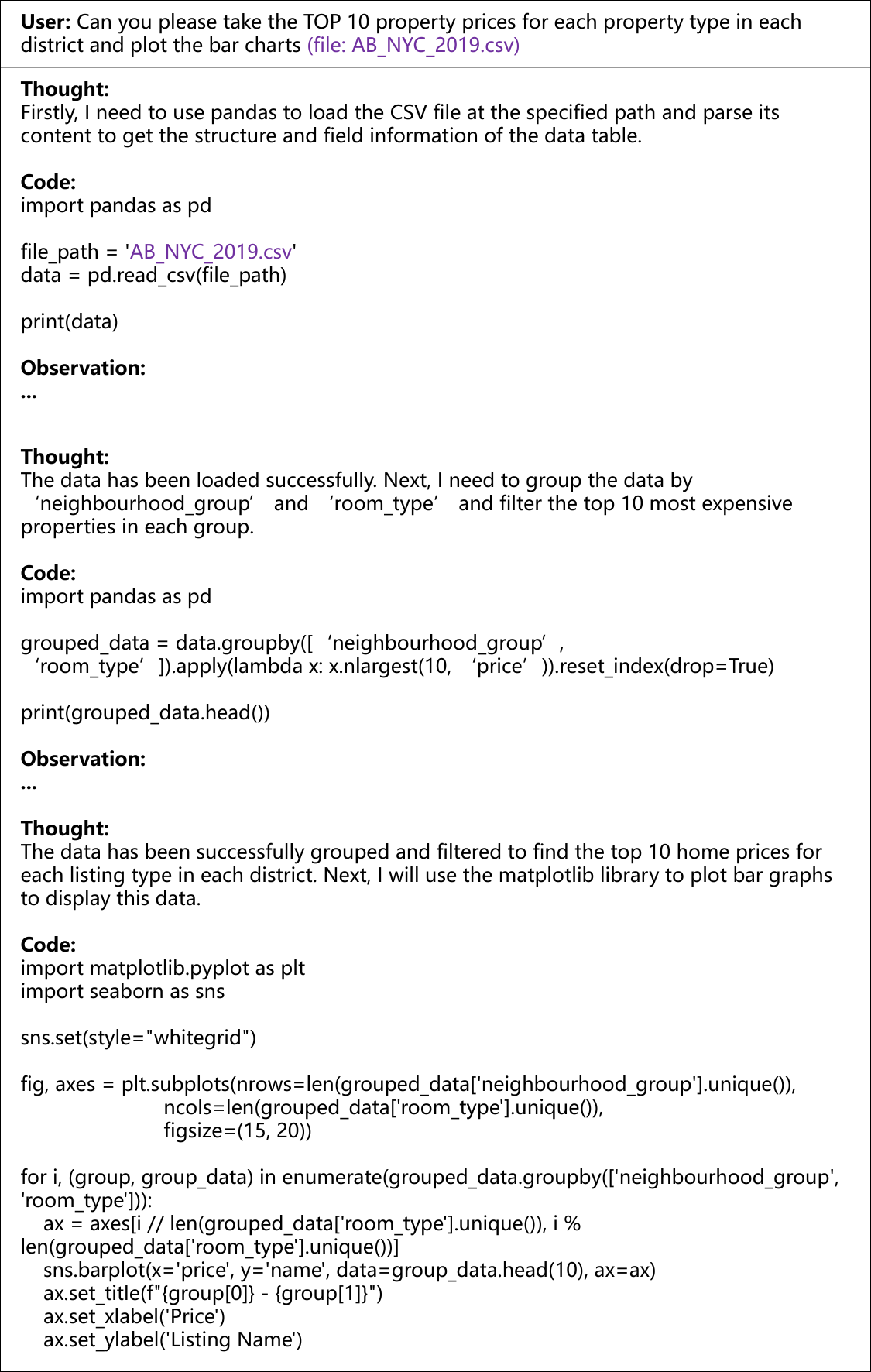} 
\end{figure*}

\section{Dataset Statistics}
\label{app:dataset}
\textbf{LegalAgentBench} LegalAgentBench includes 17 specialised databases and is equipped with 36 different tools for retrieving information from the databases. Detailed information can be found in Table \ref{tab:legalagentbench dataset}

\begin{table*}[h]
\renewcommand\arraystretch{1.1}
\resizebox{1.0\linewidth}{!}{
\begin{tabular}{cccc}
\hline
\multicolumn{1}{c}{\textbf{Corpus}} & \multicolumn{1}{c}{\textbf{Format}} & \textbf{Size} & \textbf{Brief overview}                                   \\ \hline
CompanyInfo                         & Tabular Database                    & 695           & Basic information of listed companies                     \\
CompanyRegister                     & Tabular Database                    & 10,125        & Registration information of listed companies              \\
SubcompanyInfo                      & Tabular Database                    & 9,433         & Investment information of listed companies                \\
LegalDoc                            & Tabular Database                    & 24,372        & Legal cases involving listed companies                    \\
LegalAbstract                       & Tabular Database                    & 1,200         & Summary information of legal cases                        \\
CourtInfo                           & Tabular Database                    & 3,413         & Basic information of courts                               \\
CourtCode                           & Tabular Database                    & 3,348         & Levels and administrative division codes of courts        \\
LawfirmInfo                         & Tabular Database                    & 4,768         & Basic information of law firms                            \\
LawfirmLog                          & Tabular Database                    & 101           & Service information of law firms                          \\
AddrInfo                            & Tabular Database                    & 19,533        & Province, city, and district corresponding to the address \\
RestrictionCase                     & Tabular Database                    & 46            & Cases involving restrictions on high consumption          \\
FinalizedCase                       & Tabular Database                    & 119           & Cases closed upon final enforcement                       \\
DishonestyCase                      & Tabular Database                    & 13            & Cases involving dishonest judgment debtors                \\
AdministrativeCase                  & Tabular Database                    & 443           & Cases involving administrative penalty                    \\
LegalKonwledge                      & Retrieval Corpus                    & 26,951        & Knowledge from legal books                                \\
LegalArticle                        & Retrieval Corpus                    & 55,347        & Legislatively enacted legal articles                      \\
LegalCases                          & Retrieval Corpus                    & 2,370         & Officially published guiding cases                        \\ \hline
\end{tabular}
}
\caption{}
\label{tab:legalagentbench dataset}
\end{table*}

\textbf{AgentBench} Mind2web contains 912 tasks in 73 websites covering a wide range of areas such as tourism, information and more. Database includes 1599 data entries. Database operations include select, insert and update.


\end{document}